\documentclass[final, 3p,times, authoryear]{elsarticle}

\renewcommand{\baselinestretch}{1.5} 
\usepackage{color, colortbl}
\usepackage{appendix}
\usepackage{amssymb}
\usepackage{mathtools}
\usepackage{multirow}  
\usepackage{subcaption}
\usepackage{scalefnt}
\usepackage{changepage}
\usepackage{etoolbox}
\usepackage{amsmath}
\usepackage{amssymb}
\usepackage{amsthm}
\usepackage[T1]{fontenc}
\usepackage{doi}
\usepackage[linesnumbered,ruled,vlined, noend]{algorithm2e}
\usepackage[noend]{algpseudocode}
\usepackage{float}
\usepackage{xcolor}

\DeclareMathOperator*{\argmin}{arg\,min}

\definecolor{LightGray}{gray}{0.95} 
\definecolor{Gray0}{gray}{0.9} 
\definecolor{Gray1}{gray}{0.85} 
\definecolor{Gray2}{gray}{0.8} 
\definecolor{DarkGray}{gray}{0.7} 

\begin{document}

\begin{frontmatter}

\title{Effective and interpretable dispatching rules for dynamic job shops via guided empirical learning}

\author{Cristiane Ferreira \corref{cor1}}
 \ead{cristiane.ferreira@fe.up.pt}
\author{Gon\c{c}alo Figueira}
 \ead{goncalo.figueira@fe.up.pt}
 \author{Pedro Amorim}
 \ead{pamorim@fe.up.pt}
\cortext[cor1]{Corresponding author}

\address{INESC TEC, Faculdade de Engenharia, Universidade do Porto, Rua Dr. Roberto Frias, s/n, 4600-001 Porto, Portugal}

\begin{abstract}

The emergence of Industry 4.0 is making production systems more flexible and also more dynamic. In these settings, schedules often need to be adapted in real-time by dispatching rules. Although substantial progress was made until the '90s, the performance of these rules is still rather limited. The machine learning literature is developing a variety of methods to improve them, but the resulting rules are difficult to interpret and do not generalise well for a wide range of settings.
This paper is the first major attempt at combining machine learning with domain problem reasoning for scheduling. The idea consists of using the insights obtained with the latter to guide the empirical search of the former. Our hypothesis is that this guided empirical learning process should result in dispatching rules that are effective and interpretable and which generalise well to different instance classes.
We test our approach in the classical dynamic job shop scheduling problem minimising tardiness, which is one of the most well-studied scheduling problems. Nonetheless, results suggest that our approach was able to find new state-of-the-art rules, which significantly outperform the existing literature in the vast majority of settings, from loose to tight due dates and from low utilisation conditions to congested shops. Overall, the average improvement is 19\%. Moreover, the rules are compact, interpretable, and generalise well to extreme, unseen scenarios.

\end{abstract}

\begin{keyword} Scheduling \sep Dynamic Job Shop \sep Dispatching Rules \sep Genetic Programming
\end{keyword}

\end{frontmatter}

\section{Introduction}
\label{S:introduction}

Industrial environments experience disruptive progress with the advent of Industry 4.0 (i4.0). Several technologies enable real-time interaction between physical and digital machines, such as additive manufacturing, Internet of Things, distributed ledgers, advanced robotics and artificial intelligence \citep{olsen2020}. These advancements provide more flexible production systems capable of responding quickly to new demands, breakdowns, and other unforeseen events. However, this adds also more complexity to production scheduling.
Processes are naturally more dynamic and subject to disruptions: new job arrivals, lot size changes, cancellations, production failures or machine breakdowns. The scheduling of operations must therefore be designed to respond in real-time to possible deviations \citep{parente20}. Thus, research is needed to leverage dynamic scheduling for the prompt and effective re-optimisation of operations' sequence under unexpected events. 

Different strategies are applied to manage uncertainty in dynamic scheduling. There are proactive approaches that generate robust schedules based on some prediction of the system behaviour. These methods have received some attention, but their performance highly depends on the predictive information obtained. In a hybrid predictive/reactive alternative, a baseline schedule is continuously adapted during execution. Finally, a completely reactive approach takes no decision in advance. Instead, immediate and local decisions are made in the presence of real-time events. This strategy usually adopts simple heuristics such as Dispatching Rules (DRs) to prioritise jobs waiting to be processed in a machine's queue \citep{Ouelhadj2008}. 

As simple sequencing procedures, DRs are flexible, easy to implement, and extremely fast to execute. However, their performance depends heavily on the context. \cite{Lawrence97} compare the performance of DRs and exact methods in a dynamic scheduling context. They show that DRs' performance converges to that of a dynamic and sophisticated heuristic for high uncertainty levels. Similarly, \cite{JainFoley16} conclude that under high load levels and long interruptions, the rules work better than adapting a complete schedule. Effective DRs have been developed for different problems, such as the classical job shops \citep{ChenMatis13}, shops with batch release \citep{XiongFanLi17} and parallel machines \citep{HuiPin17}. The applications are not limited to manufacturing. For instance, \cite{xu2016} approach a problem faced by subcontractors in the construction industry. \cite{JungPinedo19} use scheduling rules for operating rooms in such a way that emergency surgeries are promptly attempted. In a real case study, \cite{SweCamp19} investigate the development of rules for sequencing vessels in the Upper Mississippi River.

In this work, we aim at designing rules for the Job Shop Scheduling Problem (JSSP), one of the most studied problems in dynamic scheduling. The JSSP is defined on a set of jobs $J$ that require operations to be processed in a set of machines $M$. Each job $j \in J$ has a set operations $O_j$ that must be executed in a predefined sequence. An operation $i \in O_j$ has a specific processing time $p_{ij}$ and must be executed by a specific machine. Each machine can only process one operation at a time \citep{Pin08}.  In the dynamic version of the problem, the shop serves a continuous stream of jobs whose arrivals are usually unknown in advance. Thus, the operations' sequence must be continuously re-optimised to adapt to unexpected deviations \citep{Holthaus97}. The metric considered is the mean tardiness, which is challenging because the performance of rules highly depends on the shop load conditions, i.e., there is no rule with superior performance in both tight and light load conditions \citep{Holthaus00}. 

Dispatching rules have been traditionally derived by empirical or analytical studies, sustained by scheduling theory. To boost their effectiveness, machine learning algorithms appear as a promising alternative. For example, \cite{Shahzad12} use data-mining to analyse the behaviour of dispatching rules and generate decision trees to select them. \cite{JunLeeChun19} use Random Forests with the same purpose. There are also Reinforcement Learning approaches, using Neural Networks either to select dispatching rules or to select directly the next operation \citep{WasReiBelAlt18, Luo20}. All these approaches make use of complex, black-box models, such as Neural Networks and tree ensembles, which cannot be easily interpreted. These models are designed and validated with little domain expertise. As a result, despite their good behaviour in some tested cases, they frequently do not generalise well to unseen scenarios.  

Our approach aims at designing interpretable DRs and incorporating domain-specific knowledge. This is made possible by the use of Genetic Programming (GP), an evolutionary hyper-heuristic \citep{Burke2009}. GP evolves computer programs or mathematical expressions, which are executed to solve a problem.  The expressions might be used in different ways, such as selecting heuristics \citep{Drake20} or computing directly the decisions, like a dispatching rule. One of its advantages compared to other algorithms is the possibility of generating reasonably sized, interpretable-by-design models. By evolving models with GP, researchers might be able to understand their behaviour and even increase knowledge on the problem \citep{Kronberger10}. 

In the last decade, there has been increasing attention to the use of GP for production scheduling with promising  results \citep{Nguyen2017}. However, most papers focus on algorithmic efficiency and produce extremely long rules. Moreover, these rules work well in particular instances, but none of them performs well in a wide range of settings. We propose a guided empirical learning process, whereby insights derived from problem reasoning and existing rules are used to adjust the algorithm search space, and thus find better rules. We aim at deriving effective, compact and meaningful rules with good performance in general. To assess the generalisability of the evolved rules, we perform comprehensive computational experiments, varying due date tightness, shop utilisation and number of machines, in a wide spectrum of values. We also test the rules in settings with stochastic processing times. Our hypothesis is that by incorporating problem expertise in the algorithm configuration, it should find well-performing and more general rules, which hopefully will also be interpretable. 

The remainder of this paper is organised as follows. A literature review is provided in Section \ref{S:literature}, which first describes the studies on dispatching rules and then reports the main works on GP applied to JSSP. In Section \ref{S:methodology}, we describe our guided empirical learning approach and the algorithm implemented. We detail the learning process and the four GP variants in Section \ref{s:learning_iterations}. In Section \ref{S:results}, the evolved rules and their performance comparison are found. We conclude the paper in Section \ref{S:conclusions} with the overall achievements and some final considerations.

\section{Literature Review}
\label{S:literature}

This section describes the related work on dynamic job shop scheduling. At first, we describe the best dispatching rules manually designed with focus on the mean tardiness metric. Then we review the main achievements on the automated generation of rules using GP. 

\subsection{Dispatching rules for dynamic job shop scheduling}
\label{s:DRs_literature}

Dispatching rules are used to prioritise jobs awaiting to be processed in a queue. They assign scores to each job based on its properties or the system status. Research on the performance of these rules has a long tradition, particularly in dynamic job shops. \cite{Haupt89} and \cite{Ramasesh90} are relevant works on the topic, providing thorough explanations on the behaviour and classifications of dispatching rules. The most common performance metrics may be grouped into flowtime-related and due-date-related measures. Next, we discuss the most relevant findings on the use of dispatching rules for job shops.

The flowtime is the time the job spends in the shop from release to its completion time. In general, the SPT rule, which selects the job with the shortest operation processing time, effectively minimises mean flowtime. Although the SPT is myopic as it may select jobs that could find long queues when visiting the subsequent machines. For this reason, in addition to the current operation processing time, some rules also include \textit{look-ahead information}. \cite{Holthaus97} were the first to add the workload in the next machine queue (WINQ) to the SPT, devising the rule PT+WINQ. Later on, \cite{Rajendran99} included the processing time of the job's subsequent operation (NPT) and proposed the rule 2PT+WINQ+NPT, which remains the best for minimising the mean flowtime of jobs. 

Tardiness and percentage of tardy jobs are the most common due date related measures. Past research has shown that if due dates are loose or the shop utilisation is low, the dispatching rule should consider some due date information of jobs. 
Otherwise, in highly loaded shops, processing-time bases rules work better \citep{Blackstone82, Haupt89, Ramasesh90}. Optimising due date measures is challenging as there is no simple rule for all load conditions. However, much work has been developed in that direction. For instance, simulation studies show that prioritising \textit{operations} due date instead of \textit{jobs} due date is more effective. Operations due date are usually calculated by distributing the allowance (remaining time until due date) of jobs per remaining work (the sum of all subsequent operations processing time). 

\cite{AndersonNyiernda90} proposed the rules CR+SPT and S/RPT+SPT, that prioritise operations according to their due dates at dispatching time. To calculate due dates, while the former considers the Critical Ratio (CR), the latter uses the job slack divided by the remaining work. The CR of a job is the ratio between its allowance and its remaining work. Slack is calculated by subtracting the remaining work from the job allowance. The idea behind those rules is that the operation due date should be as short as the proportional job allowance for that operation. In practice, they apply the SPT rule if the due date is close or has passed and operations due date in the opposite case. In \cite{RaghuRajendran93}, the authors use shop utilisation as a parameter to balance between operations processing time (PT) and due date (PT$\cdot$S/RPT). They also improve the calculation of the WINQ parameter by proposing a better estimation of the next machine's queue. The resulting rule (RR) is PT$\cdot exp(u)$ + PT$\cdot$S/RPT$\cdot exp(-u)$+ WINQ, where $u$ is the mean shop utilisation. Although this rule presents good results in general, under very tight load conditions, 2PT+WINQ+NPT performs better \citep{Holthaus00}.

Instead of pursuing a unique dispatching rule, recent works focus on selecting rules according to the conditions of the production system. Very different approaches have been proposed in that direction. \cite{romero-silva18} study a flowshop with multi-class jobs. The objective is to analyse if using different dispatching rules in each queue leads to improved mean flowtime and flowtime variance. They concluded that a better performance is obtained when applying distinct rules for each machine.
In \cite{amin18}, a preference voting model is employed to select rules based on multiple criteria. The authors designed a Linear Programming Model for classifying rules according to their relative performance in each criterion. \cite{Pergher20} develop a framework for due date assignment, order release and dispatching rules according to the decision-maker preferences and multiple performance criteria. A decision support system performs an interactive process where the user inputs trade-offs judgements. 

The recent literature on dispatching rules for the JSSP has focused on specific issues, such as arrivals in batch \citep{XiongFanLi17} and precedence between jobs \citep{Fan21}. For the classical JSSP minimising the mean tardiness, the rules generated in the '90s remain the state-of-the-art. In the last ten years, the automated design of rules has received much attention, mainly with Genetic Programming. In the following, we describe the main achievements obtained with that method.

\subsection{GP for the automated design of dispatching rules}
\label{s:GP_literature}

Genetic programming has been applied in several studies on dispatching rules, with promising results in several scheduling problems. We briefly describe the main achievements for dynamic job shops in the following. For comprehensive surveys on the use of GP for production scheduling, see \cite{Branke16}, and \cite{Nguyen2017}.

\cite{Hildebrandt10} and \cite{Branke15} develop GP algorithms to minimise mean flowtime and show that their results outperform classical dispatching rules. In \cite{Nguyen15} and \cite{Nguyen18}, the authors find effective rules for several performance metrics - including mean tardiness - but they train specific rules for each load condition. To our knowledge, \cite{Karunakaran17} and \cite{ferfigamo20} are the only authors focusing on shops with stochastic processing times. However, the former do not benchmark the evolved rules with existing methods, and the latter consider deterministic job arrivals.

Some studies show that GP finds effective ensembles of heuristics for job shops. In \cite{Hart16}, the authors use GP on the design of a hyper-heuristic for the static version of the JSSP. Their heuristic applies several rules in turn and outperforms single dispatching rules. \cite{Park18} compares different combination schemes for ensembles. They show that using a linear combination of the rules' scores works better than applying single dispatching rules or ensembles with majority voting. GP may also be combined with other heuristic approaches such as Iterated Local Search \citep{Nguyen15}, or more standard evolutionary algorithms \citep{Pickardt12}.

Different authors studied methods for improving the algorithm efficiency. Surrogate models, for instance, provide faster fitness calculation. They are used to avoid employing high computational effort in the evaluation of non-promising rules.  Running these models is usually much efficient than performing a complete simulation for accessing the fitness of individuals \citep{Nguyen14, Hildebrandt15, Nguyen16}. Early studies have shown that the tree-based representation is the best approach to evolve powerful scheduling heuristics \cite{Nguyen13a}. However, as they tend to grow during the evolutionary process, alternative representation structures have also been tried \citep{Nguyen13, Branke15, Nguyen18}. Besides a good representation, an important decision consists of the terminal set, i.e., the set of the jobs and system attributes that compose the rules. \cite{Branke16} presents a detailed survey on that, and \cite{Mei16} proposes a mechanism for selecting relevant terminals.

The research on GP for the JSSP extends to variants, as the Flexible JSSP. \cite{TayHo08} were the first to evolve rules for the problem. Their work was followed by \cite{Nie13} with a Gene Expression Programming (GEP) approach. Differently from GP, GEP-based algorithms make use of a fixed-length representation structure. We also highlight the work from \cite{Zhou19}, which develops a co-evolutionary GP algorithm for due date assignment and dispatching rules. A multi-objective JSSP was tackled by \cite{Nguyen13b} and \cite{Nguyen14b}. The authors evolve rules that are non-dominated for the percentage of tardy jobs, mean and maximum tardiness, mean and maximum flowtime. Recently, \cite{Fan21} evolve dispatching rules for the JSSP with extended precedence constraints.

Despite the extensive literature on the automated development of rules for job shops, most works do not benchmark their results with state-of-the-art rules. Others train distinct rules for load condition \citep{Nguyen13b, Nguyen15, Nguyen18}. None of the previous works presents a single rule for minimising the mean tardiness that outperforms existing ones. Moreover, there is limited discussion on the size of the rules. A simplification technique is proposed in \cite{Nguyen16}, but their results are still too complex and difficult to interpret. This work aims at overcoming these gaps and find compact rules that outperform the existing ones in distinct job shops settings and load conditions.

\section{Methodology}
\label{S:methodology}

Despite the good performance of machine learning algorithms, such as Neural Networks and ensembles, most of them generate complex models. They are usually composed of constants and weights that cannot be easily interpreted. On the other hand, the optimal solutions of some optimisation problems consist of simple symbolic expressions (e.g. the Economic Order Quantity, which can be easily found by GP -- \cite{LopFigAmoLob20}). In scheduling, \cite{xu2016} show that under varying and stochastic processing costs, jobs are optimally prioritised by the \textit{Minimum Slack and Longest Remaining Work} policy. Our hypothesis is that on more complex problems, if appropriately guided, GP would still find reasonably sized and effective (if not optimal) symbolic expressions. 

In order to guide GP, we use problem knowledge in the learning process. Some initiatives with similar purpose are found in previous works, particularly in the study of hyper-heuristics. \cite{Burke2009} describes a methodology for the use of GP with contribution from domain experts. There are six steps, starting from the analysis of existing heuristics for the problem. However, human participation is limited to the initial decisions when the algorithm and function set are defined. There is no feedback from GP solutions to improve the algorithmic decisions. A more iterative process is followed by \cite{LopFigAmoLob20}.
The authors apply a cooperative co-evolutionary GP algorithm that evolves simultaneously two expressions. Each expression is assigned a specific function set, with some symbol engineering, defined based on the analysis of existing heuristics. The proposed GP finds the EOQ rule for deterministic demand scenarios and meaningful and effective expressions in the stochastic case.

\begin{figure}[ht]
  \centering
    \includegraphics[width=0.65\textwidth]{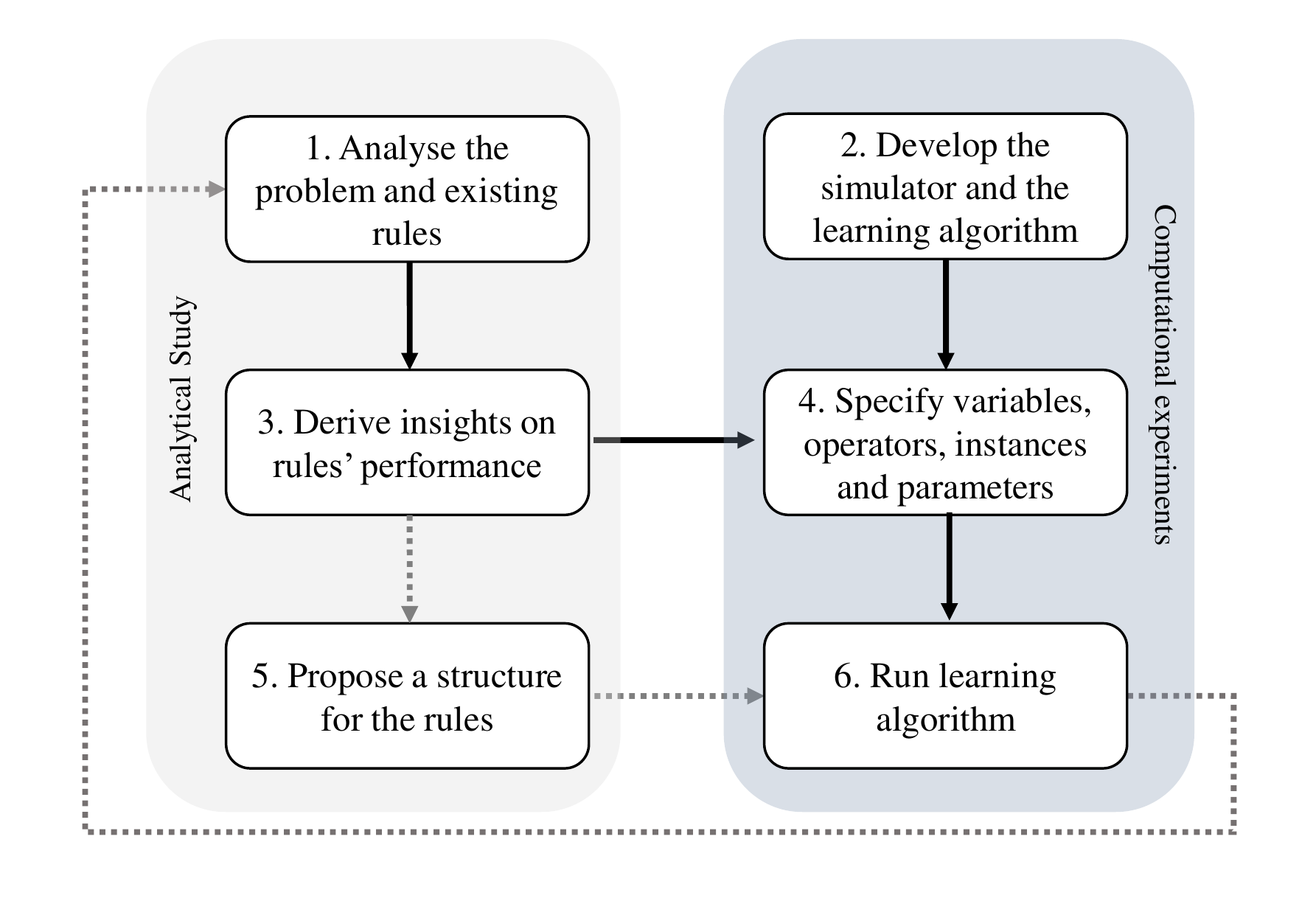}
    \caption[]{Proposed methodology, an iterative approach where reasoning on the problem and results conduct to inputs for the computational experiments. The algorithm search is combined with human intelligence.}   
    \label{fig:methodology}  
\end{figure}

In this paper, we go beyond those studies and formalise the iterative process, as illustrated in Figure \ref{fig:methodology}. Analytical reasoning on the problem (left-hand side) provides the necessary insights to specify data, parameters and constraints to boost the computational experiments (on the right). The left-hand side describes the process often found in traditional studies of dispatching rules (Section \ref{s:DRs_literature}) while the right-hand side consists of the typical methodology from the GP literature (Section \ref{s:GP_literature}).  The aim here is to cross-fertilise these two approaches into a more holistic methodology. This has resulted in the six steps of Figure \ref{fig:methodology}. Precedence relations exist between some, but not all, steps - many of them may be conducted in parallel. 

 In step 1, the problem is studied. Theoretical foundation and existing rules are analysed. This study provides conclusions on the behaviour of rules and specific terminals (step 3). There are many possible outcomes from step 3 regarding the best learning configuration, terminals to be considered, GP parameters, or constraints that must be incorporated. They may be used in step 4 to define new terminals and operators (or remove existing ones) for GP, refine the training set, or even tune the algorithm parameters. Depending on the results of step 3, one may fix a specific structure for the evolved rules and guide the learning algorithm through a restricted search space. In step 6, GP runs, generates new rules to be analysed in step 1, and a new iteration begins. Step 2 refers to the developments of the GP and the simulator, which is performed only once.  
 
 Next, in Section \ref{S:GP}, we describe the proposed GP algorithm and explain the fitness calculation, which is the average mean tardiness produced by each rule relative to a benchmark. The same metric used for training (in Section \ref{s:learning_iterations}) is applied for the validation of the results, in Section \ref{S:results}. The simulation model is detailed in Section \ref{S:simulator}, and the design of the experiments are explained in Section \ref{S:instances}. We conclude with the parameters tuning in Section \ref{S:parameters}.
 
\subsection{Learning algorithm}
\label{S:GP}

Genetic Programming (GP) evolves a population of mathematical expressions interpreted or executed to solve a problem \citep{Koza1994}. The aim is to find the highest fit individual in a search space, where each individual's fitness corresponds to how well the respective expression solves the problem. This approach uses a flexible representation scheme that, combined with a powerful search mechanism, can explore both the structure and parameters of individuals. A generic GP algorithm consists of creating an initial population that is evolved over many generations. At each generation, all individuals from the population are evaluated by the fitness function. The best ones are reproduced and combined for the next generations. The process continues until the algorithm reaches a stop criteria. 

\begin{figure}[ht]
  \centering
    \includegraphics[width=0.5\textwidth]{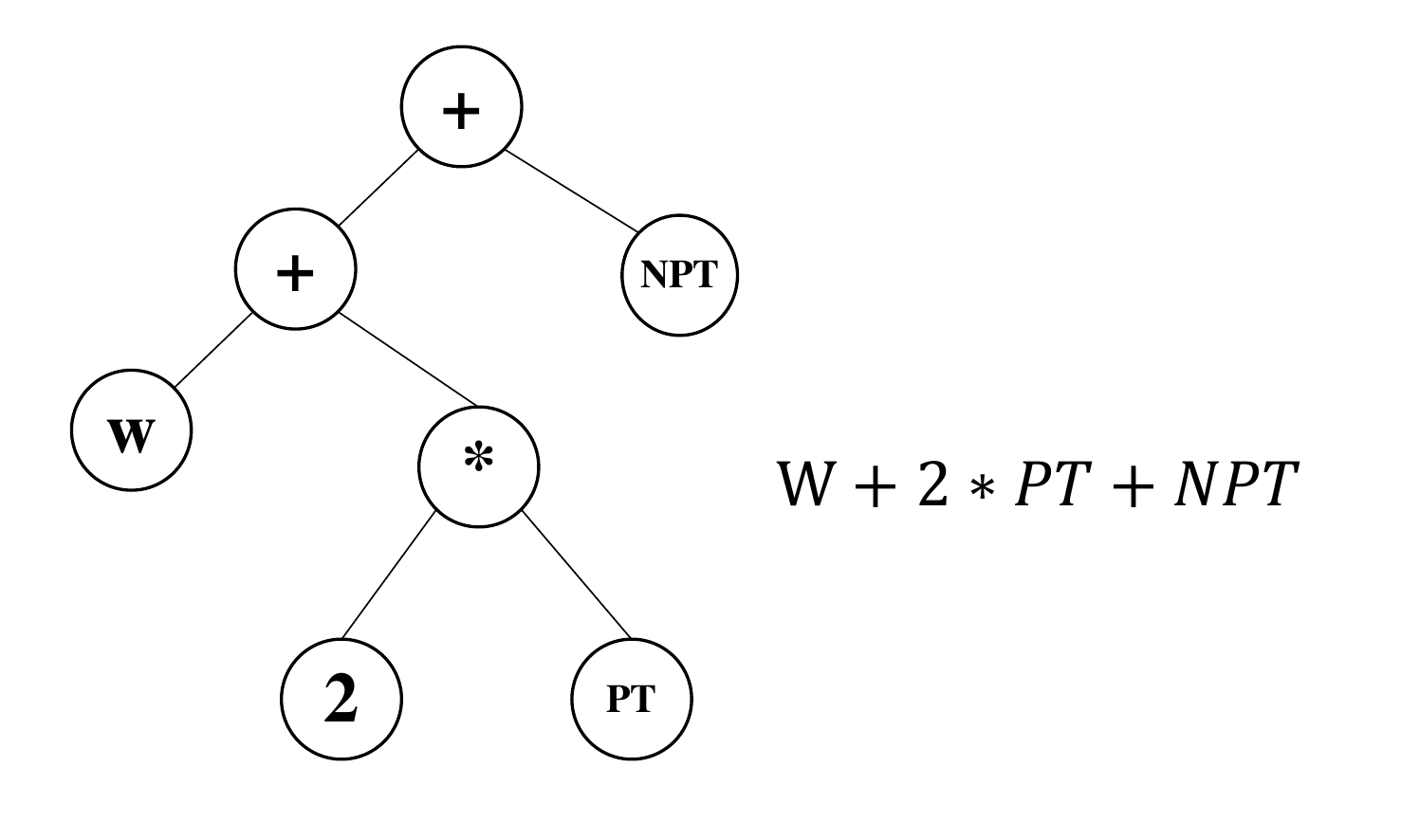}
    \caption[]{A GP individual tree example and its corresponding dispatching rule}   
    \label{fig:gp}  
\end{figure}

Symbolic expression trees are the most common representation of individuals in GP. They are  composed of a function set: \textit{terminals} and \textit{operators} appropriate to the specific problem. The former are the leaf nodes and include parameters of the problem operated by the latter. Figure \ref{fig:gp} illustrates an individual and the corresponding dispatching rule. In this example, the terminals are $2$, $PT$ (processing time of current operation), $W$ (work in the next queue) and $NPT$ (processing time of the next operation). The mathematical operators $+$ and $*$ compose the operators set. Details on the specific implementation of GP in this work are given next.

The pseudo-code of the proposed GP is provided in Algorithm \ref{alg:gp}. Each individual $A$ represents a dispatching rule whose terminals are parameters from the jobs and the shopfloor. The proposed GP runs a generational evolution where a whole population $P_g$ of individuals is evaluated and updated at each generation $g$. The generation of the initial population in line 3 uses the algorithm Ramped Half-and-Half \citep{Koza1994}, which creates random trees by combining two methods, "full" and "grow", with equal probability. The minimum and maximum depth for trees must be predefined. In trees created by the "full" method, the path length from the root to every terminal is equal to the specified maximum depth. The "grow" method instead generates trees with variable shapes and sizes.

\begin{algorithm}[ht]
\SetAlgoLined
\SetKwInOut{Input}{Input}
\Input{Training set I, benchmark values $\bar{T}_{Bi}$}
\KwResult{Dispatching rule $a^*$}
\BlankLine
\emph{$a^* \leftarrow \emptyset$}\;

\emph{$g \leftarrow 1$}\;
\emph{Initialise population $P_g$}\;

\For{$g = 1$ to $max\_g$}{

\For{$e = 1$ to $2$}{
\ForEach{individual $a \in P_g$}{
\ForEach{instance $i \in I$}{
\emph{$\bar{T}_{ai} \leftarrow$ $\frac{1}{n_e} \cdot \sum\limits_{k=1}^{n_2}{simulate(a, i)}$}\;
}
\emph{$\tau_a = \prod\limits_{i \in I }{\{\bar{T}_{ai}/\bar{T}_{Bi}\}}^{1/|I|}$}\;
\emph{$fitness(a,g) = \tau_a + |a|\cdot g \cdot \rho$}\;
}
\emph{$P_g \leftarrow$ Best 50\% individuals of $P_g$}\;
}

\emph{$a* \leftarrow \argmin\limits_{a \in P_g \cup {a*}} fitness(a, g)$}\;

\emph{$P_{g+1} \leftarrow $ Crossover and mutation in  $P_g$}\;

}
\caption{Genetic Programming algorithm}\label{alg:gp}
\end{algorithm}

The fitness computation requires running a job shop simulator, which is the most expensive procedure in the whole algorithm. Simulation studies of dynamic job shops usually apply several replicas of the same instance - each with thousands of job completions. In lines 6-10, the fitness of individual $a$ is calculated. The simulator returns the mean tardiness when applying rule $a$ for instance $i$. A reliable estimation of rules performance requires many replications of a job shop simulation. Therefore we adopt a two-step evaluation procedure. The first and lightweight step (when $e = 1$) runs a small number of replications $n_1$ for the whole population. Then, the 50\% worse individuals are discarded (line 2), and $P_g$ keeps only the most promising individuals. In the second step, a larger number of simulation replications $n_2$ is applied in the remaining population. 

An individual evaluation combines the tardiness performance and the tree size. The tardiness of a job is the difference between its completion time $c_j$ and due date $d_j$ if the job is delayed. Given a set of jobs $J$, we calculate the mean tardiness $\bar{T}$ as follows. 

$$ \bar{T} = \frac{\sum_{j \in J}{ \textrm{max}(0, c_j - d_j)}}{|J|}$$

The fitness function computes the mean tardiness performance of each individual $a$ on a training set of instances. The simulation procedure applies rule $a$ in $n_e$ replications of instance $i$, and returns the average mean tardiness $\bar{T}_{ai}$ (line 8 in Algorithm \ref{alg:gp}). We define set R = \{RR, 2PTWINQNPT, CR+SPT, S/RPT+SPT, ...\}, composed of benchmark rules mentioned in seminar works \citep{AndersonNyiernda90, RaghuRajendran93, Rajendran99, Holthaus97, Holthaus00}. $\bar{T}_{ri}$ is the mean tardiness found when running the simulator with rule $r \in R$ for a given instance $i$. The overall performance of $a$ is estimated by $\tau_a$, the mean relative performance in all training instances - the smaller, the better. The relative performance of $a$ for $i$ is calculated as follows.

$$\tau_{ai} = max(\bar{T}_{ai}, \varepsilon)/max(\min\limits_{\forall r \in R}\bar{T}_{ri}, \varepsilon)$$

Thus, the relative performance of $a$ is the ratio between the mean tardiness found when running this rule and the best mean tardiness found using existing rules. To avoid divisions by zero, we use a a small positive number ($\varepsilon = 10^{-4}$) when the value of $\min\limits_{\forall r \in R}\bar{T}_{ri}$ is null -- the same applies to $\bar{T}_{ai}$ for a fair comparison on their values. We calculate $\tau_a$ by the geometric mean (see line 9 in Algorithm \ref{alg:gp}), as it handles ratios more consistently than the arithmetic mean, i.e., given two rules $a$ and $b$, the geometric mean of $\bar{T}_{ai}/\bar{T}_{ri}$ and $\bar{T}_{ri}/\bar{T}_{ai}$ is 1, while the arithmetic mean depends on the values of $\bar{T}_{ai}$ and $\bar{T}_{ri}$. Moreover, as the training set represents very different settings, the ratio $\bar{T}_{ai}/\bar{T}_{ri}$ may scale to distinct magnitude orders. Thus, the geometric mean smooths large deviations from 1. 

In addition to relative performance, the fitness function also considers a strategy to avoid large trees. GP trees tend to grow during the search and result in complex rules - an effect known as bloat. This is a drawback of the symbolic tree representation, as complex trees are computationally inefficient and cannot be easily interpreted. Bloat control strategies are usually applied to avoid this behaviour and pursue compact trees. Our approach is to penalise the tree size in the fitness function using a bloat control factor $\rho$. The penalty depends on the number of nodes $|a|$, the number of generations completed $g$ and $\rho$. Considering the twofold objective of getting effective and small rules, the fitness of individual $a$ in generation $g$ $f(a, g)$ is calculated as follows (line 10 of Algorithm \ref{alg:gp}):

$$ f(a, g) = \tau_a + |a| \cdot g \cdot \rho$$

In line 18, the population of the next generation is created from the current population $P_g$. There are two breeding pipelines, crossover and mutation, and each one is applied according to a probability. The crossover pipeline selects two individuals as parents to produce two offspring. This operation selects a node at random in each tree and swaps the two sub-trees rooted by those nodes. Mutation replaces a sub-tree with a randomly generated one, generating a new individual. Both pipelines use tournament for parent selection, returning the best of $t$ individuals selected at random ($t$ is said tournament size). The breeding process picks certain kinds of nodes with different probabilities. For example, one can state to pick non-terminals 90\% of the time and terminals 10\% of the time. The algorithm runs until it reaches a maximum number of generations $max_g$ and returns the best individual found $a^*$.

\subsection{Simulation model}
\label{S:simulator}

The simulation model follows the literature except for the number of machines $|M|$. Most researchers consider a fixed number of ten machines \citep{Holthaus97}, but we simulate instances with distinct values. The number of operations of each job is sampled from a discrete uniform distribution whose limits depend on $|M|$. Machines are also randomly assigned to operations via a uniform distribution. Re-circulation is allowed but not in subsequent operations. We assume that operations are not preemptable, and setup times are included in operations processing time.

The simulation starts with an empty setting, and a warm-up time is considered for the system to reach a steady state. Following the typical procedure, jobs are numbered in increase sequence upon arrival. Metrics are calculated only for jobs 501 to 2500 (a total of 2000 completed jobs). The arrival of jobs follows a Poisson distribution which rate $\lambda$ is based on the utilisation level of the shop ($u$). Given the mean number of operations per job $\bar{n}$ and the mean processing time of operations $\bar{p}$, the arrival rate is calculated as follows:

$$ \lambda = \frac{u \cdot |M|}{\bar{n} \cdot \bar{p}}$$

Jobs are released to the shop when they arrive and are pushed into the first operation machines' queue. Once an operation finishes processing, the job moves to the next machine. Every time a machine is available, the dispatching rule runs and assigns a score to each job in the queue. The job with the least score is selected for execution. Given the release time $r_j$ of job $j$, its due date $d_j$ depend on the allowance factor $a$ and is calculated via Total Work Content \citep{Baker84}:

$$ d_j = r_j + a\cdot \sum_{i \in O_j}p_{ij} $$
 
\cite{Lawrence97} studied the performance of dispatching rules in job shops with uncertain processing times. Based on their work, we introduced stochastic operation times in our simulation model. Given the processing time of an operation $p_{ij}$ and a coefficient of variation (\textit{cv}), the actual processing times are found by a Gamma distribution with mean $E[p_{ij}]$ and standard deviation $\sigma_{ij} = E[p_{ij}] \times cv$. The expected processing times $E[p_{ij}]$ are the same as in the deterministic version. A pseudo-code of the simulator algorithm is found in Section \ref{S:simulator_alg} from Appendix. The next section describes the instance sets.

\subsection{Instance sets}
\label{S:instances}

An instance is defined by a tuple <$|M|$, $\bar{p}$, $a$, $u$, \textit{cv}> whose parameters are described in Table \ref{tab:instances}. As mentioned in Section \ref{S:literature}, the rules' performance depends on the load conditions, which are defined by the due date allowance ($a$) and machines utilisation ($u$). In order to consider both tight and light conditions, job shop simulation studies typically use two or three values for $a$ and $u$, in most cases varying from 3 to 8 and 85\% to 95\%, respectively \citep{Holthaus97, Rajendran99, Dominic2004, Nguyen18}. The coefficient of variation \textit{cv} determines the uncertainty level in operations processing times (usually deterministic, \textit{cv} = 0). We consider three different instance sets: training, validation and test. The training set is used during GP runs, while the validation set is used to analyse the outputs of GP and compare different versions. We use the test set to evaluate the final selected rules.

\begin{table}[ht]
\caption{Training and validation instance sets}
\centering
\begin{tabular}{c l l l l}
\hline
\textbf{Parameter} & \textbf{Description} & \textbf{Training}& \textbf{Validation} & \textbf{Test}\\
\hline
$|M|$ & Number of machines  &  10  &  10  & 10, 20, 50,  100\\
\hline
$\bar{p}$ & Mean proc. time  &  25, 50  &   25, 50, 100 & 25, 50, 100, 250, 500\\
\hline
$a$   &  Due dates allowance factor & 3, 4, 6  & 2, 3, 4, 6, 8 & 1.3, 2, 3, 4, 6, 8, 10 \\
\hline
$u$   & Shop utilisation level  & .85, .90, .97  & .80, .85, .90, .95, .97 & .70, .80, .85, .90, .95, .97, .99 \\
\hline
\textit{cv}  & Uncertainty in proc. times  & 0  & 0 & 0.0, 0.2, 0.4, 0.6, 0.8, 1.0\\
\hline
\multicolumn{2}{l}{\# Instances}  & 18  & 75 & 897\\
\hline
\end{tabular}
\label{tab:instances}
\end{table}

The training set is sufficiently diversified to assess rules in distinct load conditions but does not include extreme cases or stochastic times. This allows us to assess whether the rules can generalise, and even extrapolate, to scenarios never seen before. We simulate ten machines, and frequently used values for $\bar{p}$, $a$, and $u$. Therefore, the size of the training instance set is $1 \cdot 2 \cdot 3 \cdot 3 \cdot 1 $ = 18. As described in Algorithm \ref{alg:gp}, there are two evaluation phases. The simulator runs two replications for each training instance in the first evaluation phase, and ten in the second ($n_1 = 2, n_2 = 10$).

To represent a larger range of scenarios, the validation set is composed of five values for $a$ and $u$, resulting in 25 settings in total. As in the training set, a full-factorial combination of all values is used to compose the instance set. These values enable simulating loose due dates, where the best rule (RR, from \cite{RaghuRajendran93}) produces no tardiness and high congested shops. The shop utilisation also includes the extreme values found in previous works. For validating the rules, we ran 100 simulation replications for each instance. The number of replications is adequate to ensure that the validation of the results is statistically significant for a confidence level of 95\%. The same number of simulation replications is applied to the test set, which is explained next.

The test instance set includes, in addition to all the values of the validation set, extreme values of allowance factor (1.3 and 10) and shop utilisation (0.70 and 0.99). We aim to test the evolved rules in all load conditions found in the literature. Moreover, processing times are sampled with \textit{cv} at different levels, similarly to \cite{Lawrence97}. Previous studies show that the shop size does not affect the relative performance of the rules \citep{Blackstone82}. For that reason, almost all studies on dispatching rules for the JSSP consider ten machines or less in the simulation model. However, that may not be true for some recent and more sophisticated parameters used in certain rules. Some of them, such as WOR, the \textit{workload in the next machines' queue}, from \cite{Nguyen18}, may be influenced by the number of machines in the shop. Therefore, and to attest the generalisation of our rules, we consider larger values of $|M|$ in our test set. The number of operations is selected from U($2f$, $14f$), where $f = |M|/10$. For $|M| = 10$, the number of operations follows \cite{Holthaus00} and is selected from U(2,14). Given the large set of parameters, we do not apply a full-factorial combination on all their values on the test set. The extreme load conditions (where $a=1.3$ or $10$, $u=0.7$ or $0.99$) are tested only for $|M| = 10$, $\bar{p} = \{25, 50, 100\}$ and $cv = 0$.

\subsection{Algorithm parameters}
\label{S:parameters}

Preliminary experiments were conducted to tune the algorithm parameters. We observed that GP is not very sensitive to small changes in the crossover and mutation probabilities. Thus, we selected two extreme configurations, as shown in Table \ref{tab:crossover}, and performed 30 independent runs for each combination of crossover/mutation probability and $\rho$ (the bloat control factor). The table shows the average performance ($\tau$) of the best evolved rules, considering the validation set. We also report the average and maximum rules size, i.e., the number of nodes in the corresponding trees. Notice that, in general, the highest mutation probability leads to better performance, regardless of the bloat control factor. Moreover, for $\rho$ = 0, the smaller crossover probability produces a smaller average size. The crossover operation seems to promote unnecessarily long trees. 

\begin{table}[ht]
\caption{Performance and size of rules for distinct values of crossover and mutation rate.}
\centering
\begin{tabular}{c c c c |c c c}
\hline
& \multicolumn{3}{c|}{Crossover = 0.8/ Mutation = 0.2} & \multicolumn{3}{c}{Crossover = 0.2/Mutation = 0.8} \\
$\rho$  & Avg. $\tau$ & Avg. size & Max. size & Avg. $\tau$ & Avg. size & Max. size \\
 \hline
$10^{-4}$ & 2.46  & 12.7  & 27  &  \textbf{2.43} & 13.6  & 27 \\
$10^{-5}$ & 2.31 &  39.1 & 101 & \textbf{2.06}  & 40.2  & 73\\
0 & 2.39  & 197.8  &  389  & \textbf{2.05} & 98.6  & 207 \\
\hline
\end{tabular}
\label{tab:crossover}
\end{table}

The GP parameters used in the final experiments are found in Table \ref{tab:GPParameters}. Limiting the tree depth to small values prematurely prunes good branches, so we decided to set the maximum depth of trees to a large number commonly used (17). As the bloat control factor $\rho$ impacts both the size of the rules and their performance, it received special attention. Section \ref{s:learning_iterations} details the results obtained for different values of $\rho$. We observed that the larger the population size, the better the average results obtained, although for values above 200, the improvement in solution quality is negligible. The number of generations ($max\_g$) is limited to 50, as in preliminary runs, no improvement in the fitness function was observed beyond generation 35. As the population is usually too homogeneous after some generations, the tournament size was set to the smaller possible value (2) to promote diversity. We set the values for the probability of selecting a terminal and a non-terminal (0.1 and 0.9, respectively) and the parameters of the Ramped Half-and-Half algorithm (minimum tree depth of 2 and maximum 6). All algorithms were coded in Java, and the GP procedure uses the evolutionary computation library ECJ \citep{Luke17}.

\begin{table}[ht]
\caption{GP Parameters Setting}
\centering
\begin{tabular}{l  l}
\hline
Population size & 200 \\
\hline
Number of generations  ($max\_g$) & 50 \\
\hline
Tournament size ($t$) & 2 \\
\hline
Maximum trees depth & 17 \\
\hline
Crossover probability & 0.2 \\
\hline
Mutation probability & 0.8 \\
\hline
\end{tabular}
\label{tab:GPParameters}
\end{table}

\section{Learning iterations}
\label{s:learning_iterations}

We performed four iterations on the procedure described in Figure \ref{fig:methodology}, which have resulted in four GP variants: \textit{GPLit}, \textit{GPBasic}, \textit{GPImp} and \textit{GPStruct}. The first two variants are based on the set of terminals used by the GP and scheduling literature, respectively. In the last two iterations, we improve the function set and propose a specific structure for the rules based on the analysis of the problem and the results obtained. The following sections describe each variant and the results obtained with them. 

We present beforehand the list of terminals of GPBasic, GPImp and GPStruct in Table \ref{tab:terminals}. The second column of the table shows the size associated with each terminal, i.e., the number of atomic parameters and operators. CR, for instance, has size three, corresponding to AJ/RPT (``AJ'', ``/'', ``RPT'') and PT$\cdot$CR has size five, corresponding to ``PT'', ``$\cdot$'', ``AJ'', ``/'' and ``RPT''. To calculate WINQ (the work in the next machines' queue), we use the procedure described in \cite{RaghuRajendran93}. As the terminal set of GPLit is too extensive, it is presented in Section \ref{app:terminals} of Appendix. 

\renewcommand{\baselinestretch}{1.3} 
\begin{table}[ht]
\scalefont{0.8}
\caption{Terminal sets considered in GPBasic, GPImp and GPStruct}
\centering
\begin{tabular}{l  c l c c c c}
\hline
\textbf{Terminal} & \textbf{Size} & \textbf{Description} &  \textbf{GPBasic} &  \textbf{GPImp} &  \multicolumn{2}{c}{\textbf{GPStruct}} \\
&  &  &  & & $Slack > 0$  & $Slack \leq 0$ \\
\hline
PT & 1 & Processing time of current operation  & x & x & x & x\\
\hline
WINQ & 1 & Work in the next queue  & x & x  & x& x\\
\hline
NPT & 1 & Processing time of next operation   & x & & & \\
\hline
U & 1 & Shop utilisation & x &  & & \\
\hline
PT$\cdot$S$/$RPT & 5 & Operation due date (using slack) & x &  & & \\
\hline
PT$\cdot$CR & 5 & Operation due date (using allowance) &x  &  &  & \\
\hline
S$^+$ & 1 & Slack  if positive  &  & x & x & \\
\hline
AJ$^+$ & 1 & Allowance  if positive  &  & x & x & \\
\hline
S$^+/$RPT & 3 & Relative slack if positive  &  & x & x & \\
\hline
CR$^+$ & 3 & Critical ratio if positive &  & x & x & \\
\hline
PT$\cdot$S$^+/$RPT & 5 & Operation due date (using slack) if positive &  & x & x& \\
\hline
PT$\cdot$CR$^+$ & 5 &  Operation due date (using allowance) if positive &  & x & x& \\
\hline
RPT & 1 & Remaining processing time  &  & x & x& \\
\hline
\end{tabular}
\label{tab:terminals}
\end{table}
\renewcommand{\baselinestretch}{1.5} 

\subsection{Using terminals from the literature}
\label{s:gp1}

In the first two iterations, the function sets follow the literature in two different ways. While the first (GPLit) reproduces previous GP experiments, the second (GPBasic) is based on the classical dispatching rules. Next, we detail both sets and the average results obtained with them. 

We provide two terminal sets for GPLit, namely GPLit$_a$ from \cite{Nguyen13b} and GPLit$_b$ based on \cite{Nguyen18}. In both works, GP finds well-performing rules for specific job shop settings. Ephemeral random constants are not included in our experiments to promote interpretability and generalisability. In the operators set, we use the basic mathematical operations (+, -, $\cdot$, /), \textit{max} and \textit{min}, as in the works mentioned. GPLit$_b$ also contains an \textit{If} operator, as in \cite{Nguyen18}. In all experiments the protected division applies, i.e., $a / b = 1$ if $b = 0$.

In the second iteration, we aim at providing GP with a smaller set of selected terminals and operators. The earlier studies on dynamic job shop scheduling show that look-ahead parameters should be incorporated into the SPT rule. Moreover, due date information (slack or allowance) must be considered to prioritise the most urgent operations. These parameters are found in the best existing rules, such as RR, 2PT+WiNQ+NPT, CR+SPT and S/RPT+SPT. Thus, the terminal set of GPBasic contains only the parameters present in those rules, including the processing time of current operation (PT), due date information (S/RPT and CR), next operation processing time (NPT) and work in the next queue (WINQ). The parameters $exp(u)$ and $exp(-u)$ from RR are omitted since there is no justification for using the exponential function. GPBasic and the next two variants use only the basic four mathematical operations (+,-,$\cdot$ and /) in the operators set.

\begin{figure*}[ht]
\begin{center}
\begin{subfigure}{0.32\textwidth}
	\includegraphics[width=\textwidth]{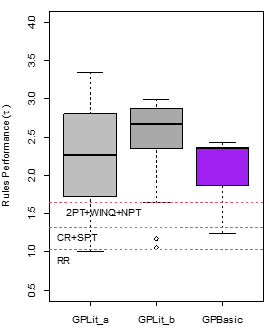}
	\caption{ $\rho = 10^{-4}$} \label{fig:4a}
\end{subfigure}	
\begin{subfigure}{0.32\textwidth}
	\includegraphics[width=\textwidth]{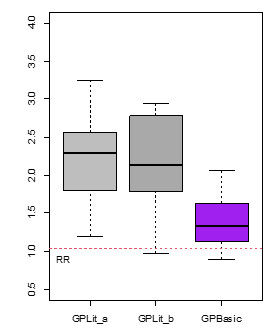}
	\caption{ $\rho = 10^{-5}$} \label{fig:4b}
\end{subfigure}	
\begin{subfigure}{0.32\textwidth}
	\includegraphics[width=\textwidth]{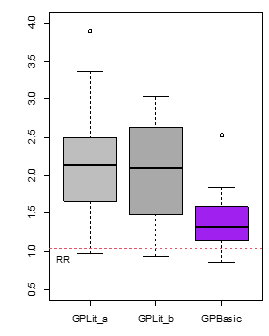}
	\caption{ $\rho = 0$} \label{fig:4c}
\end{subfigure}	
\caption{Performance of rules evolved with GPLit$_a$, GPLit$_b$ and GPBasic}
\label{fig:rulesPerformance1}
\end{center}
\end{figure*}

We describe the results of these two variants in the following. The rules' performance is calculated using the metric for training described in Section \ref{S:GP}. We set four distinct values for the bloat control factor ($\rho = \{10^{-3}, 10^{-4}, 10^{-5}, 0\}$) to explore the trade-off between rules size and performance. Following \cite{Nguyen18}, 30 independent executions were performed for each GP variant and value of $\rho$. When $\rho = 10^{-3}$, both versions of GPLit get stuck in tiny rules with poor performance ($\tau \sim 3$). Similarly, GPBasic produces PT+WINQ, which is not effective for mean tardiness ($\tau = 2.36$) as it does not consider due dates. The for the other values of $\rho$ is presented in Figure \ref{fig:rulesPerformance1}. The average performance of existing rules is also indicated by horizontal dashed lines. 

The overall performances of GPLit$_a$ and GPLit$_b$ are similar. In 90 runs, GPLit$_a$ resulted in four good rules ($\tau < 1$) and GPLit$_b$ in only two. Between both versions, GPLit$_b$ found the best rule ($\tau$ = 0.93) and its average performance is slightly better, 2.24 against 2.34 from GPLit$_a$. The best rules are found when no bloat control is applied or using light bloat control. Regarding GPBasic, when $\rho = 10^{-4}$, the algorithm generates PT+WINQ in 17 out of 30 runs, resulting in poor average performance. However, for $\rho > 10^{-4}$, this version performs better than GPLit$_a$ and GPLit$_b$. Nine rules are better than RR, and the best performance is 0.86. From these results, we conclude that with the same parameters, GP was able to find a better combination than manually devised rules. Moreover, we suspect that the function sets of GPLit$_a$ and GPLit$_b$ are probably unnecessarily large. As a consequence, there is no dimensional consistency in the rules devised with them - some terminals refer to the number of operations, others to the remaining work or even the number of machines yet to be visited. Combining these terminals is thus an additional challenge for GP. It is also interesting to observe that terminal U, which is used to adapt rules to the shop utilisation in the rule RR, is not present in the best evolved rules with GPBasic.

\begin{figure*}[ht]
\begin{center}
\begin{subfigure}{0.32\textwidth}
	\includegraphics[width=\textwidth]{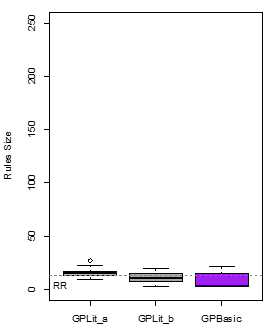}
	\caption{ $\rho = 10^{-4}$} \label{fig:5a}
\end{subfigure}	
\begin{subfigure}{0.32\textwidth}
	\includegraphics[width=\textwidth]{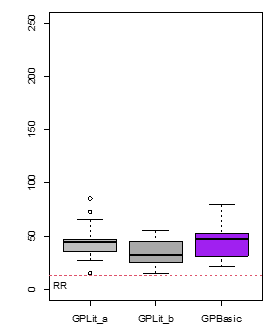}
	\caption{ $\rho = 10^{-5}$} \label{fig:5b}
\end{subfigure}	
\begin{subfigure}{0.32\textwidth}
	\includegraphics[width=\textwidth]{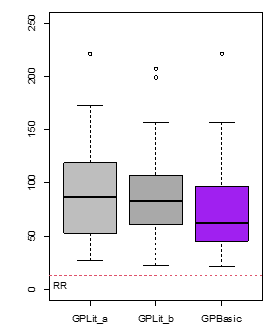}
	\caption{ $\rho = 0$} \label{fig:5c}
\end{subfigure}	
\caption{Size of rules evolved with GPLit$_a$, GPLit$_b$ and GPBasic} 
\label{fig:rulesSize}
\end{center}
\end{figure*}

Figure \ref{fig:rulesSize} presents the size of all rules evolved in these two first iterations and the size of RR for reference. We observe that our bloat control approach is effective to provide smaller rules, although the average performance is similar for $\rho = 10^{-5}$ and $\rho = 0$ (see Figure \ref{fig:rulesPerformance1}).  The good performance of GPBasic obtained at the cost of considerably longer rules. Indeed, there is no significant difference in the size obtained with the different GP variants. These first experiments demonstrate that GP can produce rules that perform well on average, but there is room for improvement. The well-performing rules are too long and cannot be easily interpreted. In the next section, we detail how the terminal set of GPBasic is improved in the last two iterations. 

\subsection{Improving the terminal set}
\label{s:gp3}

Highly loaded settings and tight due dates lead to negative values for allowance and slack, which introduces bias in the rules' behaviour \citep{Haupt89, AndersonNyiernda90}. The term CR$\cdot$PT prioritise operations with long instead of short processing times if CR is negative. Moreover, these parameters are not relevant in tight load conditions, even for optimising due date measures \citep{Ramasesh90}. Thus, in GPImp, the slack and allowance of jobs are considered only when positive; we use 0 otherwise. Terminal S (slack) is substituted by S$^+$ (slack if positive), and  CR becomes CR$^+$ (critical ratio if positive). Hence, GPImp uses PT$\cdot$S$^+/$RPT instead of PT$\cdot$S$/$RPT and PT$\cdot$CR$^+$ instead of PT$\cdot$CR as well. We also decompose some terminals into their atomic parameters to verify if GP finds better combinations. For instance, in addition to PT$\cdot$S$^+/$RPT, we also include PT, S$^+$, RPT and S$^+/$RPT. Terminals NPT (next operation processing time) and U (shop utilisation) are discarded, as they do not appear in the best rules evolved with GPBasic. 
	
\begin{figure}[ht]
  \centering
    \includegraphics[width=0.5\textwidth]{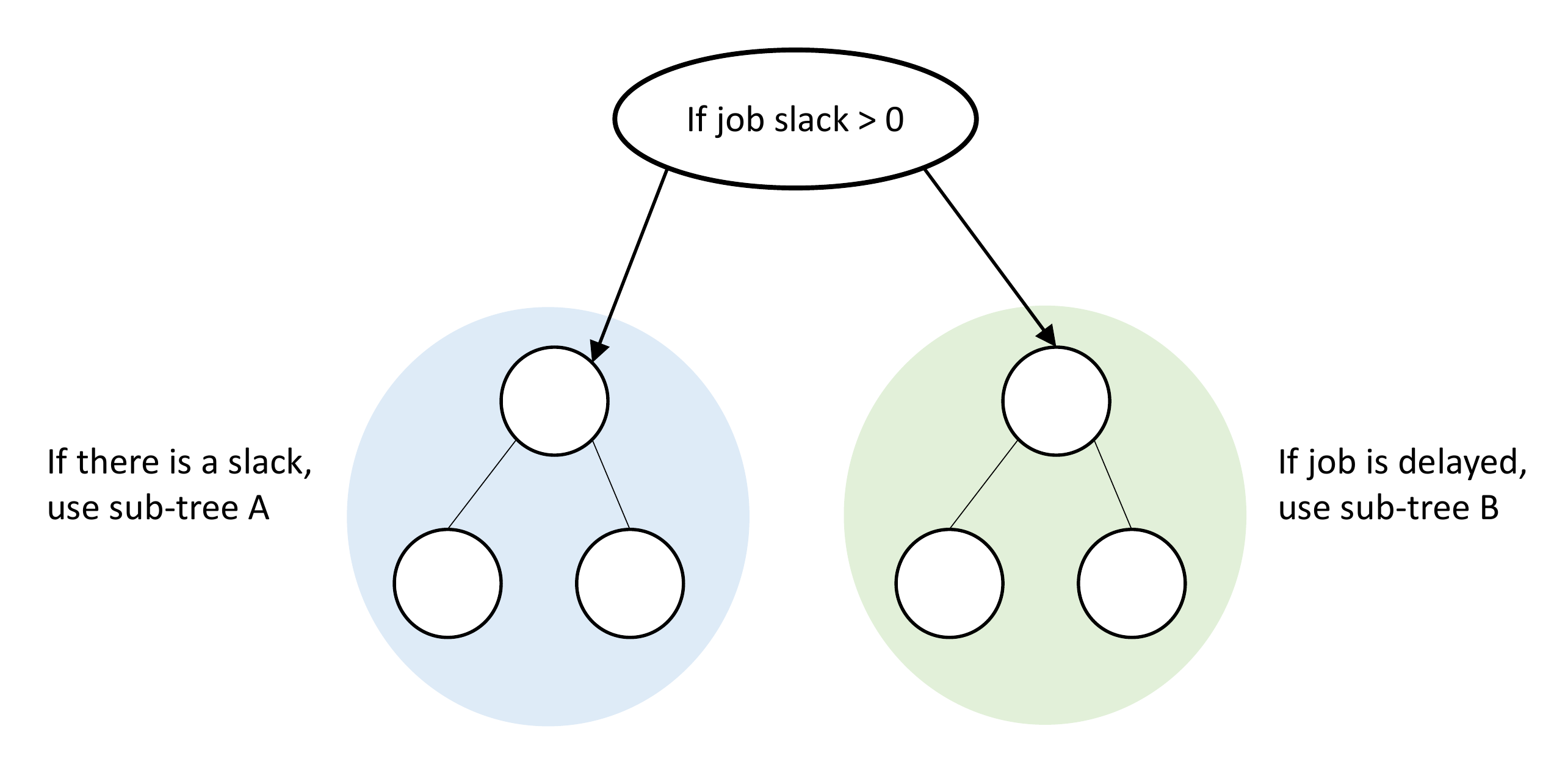}
    \caption[]{Fixed structure proposed for rules evolved}   
    \label{fig:gp4}  
\end{figure}

The last learning iteration designs a preliminary structure for the rules. If job slack is negative, it is not possible to meet the due date. From the good performance of the rules generated with GPImp, we conclude that it is better to not consider any due date information in such cases. Thus, in GPStruct, two expressions are used to calculate the score of each operation, depending on job slack. The idea is similar to CR+SPT and S/RPT+SPT, which apply operations processing time if the due date is too close or has passed. GPStruct evolves two separated sub-trees, A and B, fed with distinct terminal sets, as specified in Figure \ref{fig:gp4}. While A uses the same set of GPImp, B does not consider any due date-related terminal (as slack or allowance). Although breeding is performed independently on each sub-tree, the fitness calculation uses the whole tree. The last two columns of Table \ref{tab:terminals} indicate the terminals of GPStruct in both cases.  

\begin{figure*}[ht]
\begin{center}
\begin{subfigure}{0.32\textwidth}
	\includegraphics[width=\textwidth]{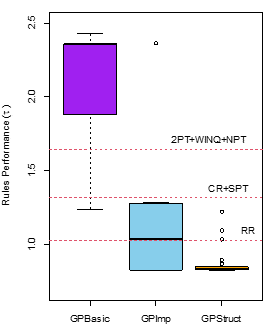}
	\caption{ $\rho = 10^{-4}$} \label{fig:6a}
\end{subfigure}	
\begin{subfigure}{0.32\textwidth}
	\includegraphics[width=\textwidth]{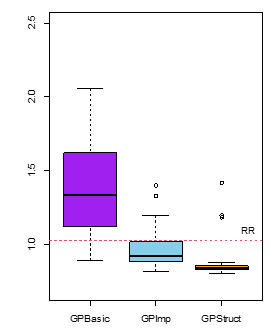}
	\caption{ $\rho = 10^{-5}$} \label{fig:6b}
\end{subfigure}	
\begin{subfigure}{0.32\textwidth}
	\includegraphics[width=\textwidth]{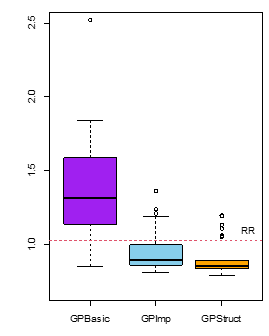}
	\caption{ $\rho = 0$} \label{fig:6c}
\end{subfigure}	
\caption{Performance of rules evolved with different GP configurations} \label{fig:rulesPerformance2}
\end{center}
\end{figure*}

Figure \ref{fig:rulesPerformance2} presents the rules' performance by GP variant - GPLit$_a$ and GPLit$_b$ are not included given their poor performance. GPImp provides much better rules than GPBasic, and is in turn outperformed by GPStruct. GPStruct provides the best performing rule ($\tau$ = 0.80) and the best average performance. Moreover, GPStruct consistently generates good rules, regardless the value of $\rho$ - $\tau \leq 0.86$ for 79\% of the rules evolved. For instance, the rule \textit{If (slack > 0)} PT$\cdot$CR$^+$ + 2WINQ \textit{Else} 2PT + WINQ ($\tau = 0.84$) was generated in 56\% of the runs with $\rho = 10^{-4}$. It is interesting that when PT$\cdot$CR$^+$ is not used, GP finds the same expression from the combination of the nodes \{PT, ``$\cdot$'', CR$^+$\}. 

Table \ref{tab:terminal_sets} summarises the average performance and the best rule of GPLit (joint GPLit$_a$ and GPLit$_b$), GPBasic, GPImp and GPStruct. The improvement in performance provided by each iteration is clear, regardless of the bloat control factor ($\rho$). On average, the rules evolved with GPBasic present a performance of 1.61 while GPImp and GPStruct find rules with 1.05 and 0.89, respectively. We also report the average size of the rules. As the rules evolved with GPStruct combine two expressions, they are slightly greater than the others.

 \renewcommand{\baselinestretch}{1.3} 
\begin{table}[ht]
\scalefont{0.8}
\caption{Performance and size of rules evolved using preliminary versions of GP. }
\centering
\begin{tabular}{c c c c |c c c|c c c |c c c}
\hline
\multirow{2}{*}{$\rho$}  & \multicolumn{3}{c}{\textbf{GPLit}} & \multicolumn{3}{c}{\textbf{GPBasic}} & \multicolumn{3}{c}{\textbf{GPImp}} & \multicolumn{3}{c}{\textbf{GPStruct}}\\
\cline{2-13}
 & $\tau$ & $\tau_{min}$ $\tau$ & Size & $\tau$ & $\tau_{min}$ $\tau$ & Size & $\tau$ & $\tau_{min}$ $\tau$ & Size &  $\tau$ & $\tau_{min}$ $\tau$ & Size \\
 \hline
\textbf{$10^{-4}$} & \cellcolor{DarkGray}2.40 &\cellcolor{Gray1} 1.00 & 13.6 &\cellcolor{DarkGray}  2.07 &\cellcolor{DarkGray} 1.24 & 8.3 &\cellcolor{DarkGray} 1.23 & \cellcolor{Gray0}0.83 & 12.0 & \cellcolor{Gray0}  \textbf{0.87} &\cellcolor{Gray0}  0.83 &15.7  \\
\textbf{$10^{-5}$} &\cellcolor{DarkGray} 2.20 & \cellcolor{Gray1} 0.97 & 39.0 & \cellcolor{DarkGray} 1.39 & \cellcolor{Gray0} 0.82 & 45.7 &\cellcolor{Gray1}   0.97 & \cellcolor{Gray0}0.81 & 47.0 &\cellcolor{Gray0}   \textbf{0.89}&\cellcolor{Gray0} 0.80 & 34.8 \\
\textbf{$0$} &\cellcolor{DarkGray} 2.11  & \cellcolor{Gray0} 0.93 & 91.7 &\cellcolor{DarkGray} 1.36 & \cellcolor{Gray0} 0.86 & 76.3 &\cellcolor{Gray1} 0.95 & \cellcolor{Gray0}0.81 & 76.3 &\cellcolor{Gray0}   \textbf{0.91} &\cellcolor{Gray0} 0.80 & 114.3 \\
\hline
\textbf{Avg.}& \cellcolor{DarkGray}2.24 &\cellcolor{Gray1} 0.97 & 48.1 &\cellcolor{DarkGray} 1.61 & \cellcolor{DarkGray} 1.00 & 43.4 &\cellcolor{DarkGray} 1.05&\cellcolor{Gray0} 0.81 & 45.1 &\cellcolor{Gray0}  \textbf{0.89} &\cellcolor{Gray0} 0.81 & 54.9 \\
\hline
\end{tabular}
\label{tab:terminal_sets}
\end{table}
 \renewcommand{\baselinestretch}{1.5} 

\section{Evolved Rules}
\label{S:results}

In this section, we exhibit and analyse the best evolved rules considering a two-criteria objective, finding \textit{effective} and \textit{interpretable} rules. These objectives are measured by the rules' relative performance in the test set and their size, respectively. For presenting the detailed results in the test set, we select non-dominated rules in both criteria. Next, we explain this selection.

\subsection{Rules selection}

\begin{figure*}[ht]
\begin{center}
	\includegraphics[width=0.9\textwidth]{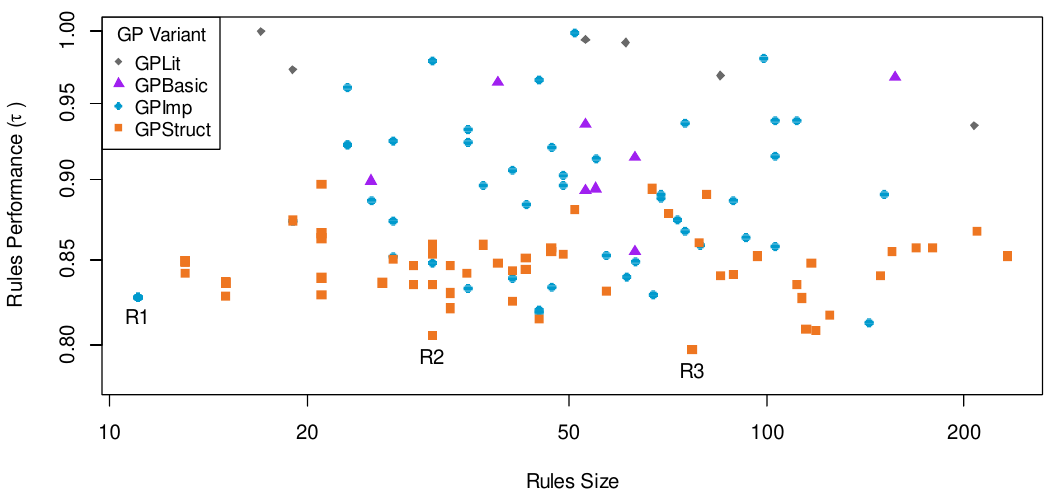}
\caption{Size and relative performance of the evolved rules. }
\label{fig:size_performance}
\end{center}
\end{figure*}

The chart in Figure \ref{fig:size_performance} presents the performance on the validation set ( \textit{y-axis}) and size (\textit{x-axis}) of all evolved rules with $\tau \leq 1$. Both axes are on a logarithmic scale. Observe that the rules generated with GPLit  (GPLit$_a$ and GPLit$_b$) are all dominated. The chart highlights three non-dominated rules (R1, R2, R3), evolved with GPImp and GPStruct. Among them, R1 is the shortest and least performing (0.83). Rule R3 is the largest and best performing (0.80), although R2 presents a very similar performance (of around 0.80).

Table \ref{tab:rules} presents the mathematical expression of each non-dominated rule. R2 was simplified by applying mathematical operations, e.g., $a + a + a$ is converted to $3a$, and $a / a$ is 1. Column ``Size'' shows the size of the expression after these modifications and the original size in brackets. R3 is too large and cannot be easily interpreted. Thus we do not present it. We detail R1 and R2 next.

\begin{table}[ht]
\caption{Evolved rules}
\centering
\begin{tabular}{l c c c c}
 \hline
\textbf{Rule} &  \textbf{Mathematical expression} &  \textbf{Size} & $\tau$ &  \textbf{Algorithm}\\
\hline
 R1 & 2PT + PT$\cdot$S$^+/$RPT + WINQ  & 11 & 0.828 & GPImp \\
\hline
\multirow{4}{*}{R2} &\textit{If (Slack > 0) Then}  & \multirow{4}{*}{19 (31)} & \multirow{4}{*}{0.805} & \multirow{4}{*}{GPStruct}   \\
  & PT + PT$\cdot$CR$^+$ + WINQ  &  & &\\
  & \textit{Else}  &  & & \\
 & 3PT + WINQ - (WINQ$/PT$) &  & & \\
\hline
R3 & (large expression)  & 77 & 0.797 & GPStruct \\
\hline
\end{tabular}
\label{tab:rules}
\end{table}

R1 has a structure similar to RR from \cite{RaghuRajendran93}. Both rules are composed of three terms, PT (operations processing time), PT$\cdot$S$/$RPT (operations due date) and WINQ (work in the next machines' queue). The difference is in the balance between PT and PT$\cdot$S$/$RPT. The rule RR use weights $exp(u)$ and $exp(-u)$ for PT and PT$\cdot$S$/$RPT, respectively, adapting the influence of those terms to the shop utilisation $u$. R1, instead, omits the term PT$\cdot$S$/$RPT for delayed jobs and applies weight 2 for PT (in RR, the weight of PT varies from 2.0 to 2.7). For negative or null slacks, R1 behaves just like 2PT + WINQ, which is an expression frequently found by GPStruct for the branch of non-positive slacks. 

The rule R2 is equivalent to R1 if the job slack is positive. The terminal CR corresponds to AJ/RPT. As slack (S) is AJ - RPT, CR = (S+RPT)/RPT = 1 + S/RPT. Thus, PT$\cdot$(2 + S/RPT), from R1, corresponds to PT$\cdot$(1 + CR), from R2. The behaviour of both rules is different only when the slack is negative or zero. In that case, R2 assigns weight 3 for PT and subtract WINQ/PT, favouring jobs with short processing times. The balance between operations processing time (PT) and work in the next queue (WINQ) in R2 is 3:1, similar to 2PT+WINQ+NPT - the best existing rule for extreme tight load settings. Except when PT is considerably small, the term WINQ/PT is much smaller than the other terms in the expression, with almost no influence on the resulting score. Thus, R2 emphasise the use of the shortest processing time for jobs in congested shops.

\subsection{Performance under deterministic processing times}

This section details the results obtained with R1 and R2 in our test set for deterministic processing times. We recall that the performance $\tau_a$ is the mean relative performance of rule $a$ against the best existing rules (described in Section \ref{s:DRs_literature}). For each instance $i$, we calculate $\bar{T}_{ai}/\bar{T}_{ri}$, where $r$ is the best rule for $i$ among all of the benchmarked rules. Thus, if $\tau_a < 1$, rule $a$ outperforms, on average, all existing rules combined. 

Table \ref{tab:performance_det1} presents the value of $\tau$ for each combination of shop utilisation and allowance factor. In these experiments we use the most common values for processing times ($\bar{p} = \{25, 50, 100\}$) and number of machines ($|M| = 10$). We performed statistical tests to compare the result of each evolved rule to the best existing rules. We obtained significant results at the 0.05 level for almost all settings -- the exceptions are marked with \dag. In these cells, we consider that there is a tie between the evolved rule and the benchmark as in cells with $\tau = 1.00$. 

The cells in the top right corner correspond to light load settings - large allowance factor and low shop utilisation. In these settings, the (evolved and benchmark) rules usually produce no tardiness, resulting in $\tau = 1.00$ (tie). Considering all settings, R1 and R2 perform better than the benchmark. R1 outperforms existing rules in 33 out of 49 settings and is worse in 7. R2 has better average performance than R1 and works better in a larger number of settings, 38 out of 49. The effectiveness of both rules is due to a better balance between operations processing time and due date, especially when job slack is negative. However, the evolved rules deteriorate under higher utilisation levels. Observe that $\tau > 1$ more frequently in cells where $u \geq$ 0.97. As the average queue size increases with utilisation level, the value of WINQ becomes larger, but the weight of PT does not follow. We conclude that it could be desirable to train specific rules for these tight load settings. 
 
 \renewcommand{\baselinestretch}{1.3} 
 \begin{table}[H]
 \scalefont{0.8}
\caption{Performance under deterministic processing times and distinct values of shop utilisation and allowance factor.}
\centering
\begin{tabular}{c c c c c c c c c | c c c c c c c c }
\hline
 & \multicolumn{8}{c|}{\textbf{R1}} & \multicolumn{8}{c}{\textbf{R2}} \\
$u \ a$ & 1.3 &  2 & 3&  4 &6 &  8 &  10  & \textbf{Avg} &  1.3 &  2 &   3&  4 &  6 & 8 &  10  & \textbf{Avg} \\
\hline
0.70 &\cellcolor{Gray1}  0.98&	\cellcolor{Gray0}0.94&	0.78&	0.53&\cellcolor{Gray1} 	 1.00  &\cellcolor{Gray1} 	1.00&\cellcolor{Gray1} 	1.00 & \cellcolor{Gray0} \textbf{0.87} & \cellcolor{Gray1}  0.98&\cellcolor{Gray0}	0.94&	0.78&	0.56&\cellcolor{Gray1} 	1.00 &	0.79 &\cellcolor{Gray1} 	1.00 & \cellcolor{Gray0} \textbf{0.85}\\

0.80 & \cellcolor{Gray1} 0.98	&\cellcolor{Gray0}  0.94& \cellcolor{Gray0}	0.87&	0.73&	0.42&\cellcolor{Gray1} 	1.00 & \cellcolor{Gray1} 	1.00 & \cellcolor{Gray0} \textbf{0.82}  &\cellcolor{Gray1}  0.99 &\cellcolor{Gray0}	0.94 & \cellcolor{Gray0}	0.88&	0.73&	0.41&\cellcolor{Gray1} 	1.00&	0.46& \textbf{0.73}\\

 0.85 & \cellcolor{Gray1}  0.99&\cellcolor{Gray1} 	0.97& \cellcolor{Gray0}	0.85& \cellcolor{Gray0}	0.82&	0.56&	0.30&	\cellcolor{Gray1} 1.00& \textbf{0.73} &\cellcolor{Gray1}   0.99&\cellcolor{Gray1} 	0.96& \cellcolor{Gray0}	0.83& \cellcolor{Gray0}	0.81&	0.57&	0.26&\cellcolor{Gray1} 	1.00 & \textbf{0.72}\\

 0.90  & \cellcolor{Gray1}  1.00 &\cellcolor{Gray1} 	0.99 &\cellcolor{Gray0}	0.90&\cellcolor{Gray0}	0.86&\cellcolor{Gray0}	0.90&	0.69&	0.07&  \textbf{0.62} & \cellcolor{Gray1}  0.98&\cellcolor{Gray1} 	0.98& \cellcolor{Gray0}	0.89& \cellcolor{Gray0}	0.85& \cellcolor{Gray0}	0.86&	0.62&	0.25& \textbf{0.72} \\

 0.95  & \cellcolor{Gray1}   0.98&\cellcolor{Gray1} 	0.98  &\cellcolor{Gray1} 	0.96& \cellcolor{Gray0}	0.86&\cellcolor{Gray0}	0.91&\cellcolor{Gray1} 	0.95 & \cellcolor{Gray0}	0.81  & \cellcolor{Gray0} \textbf{0.92} &\cellcolor{Gray1}  0.98&\cellcolor{Gray1} 	0.97&\cellcolor{Gray0}	0.93& \cellcolor{Gray0}	0.85&\cellcolor{Gray0}	0.90& \cellcolor{Gray0}	0.89& \cellcolor{Gray0}	0.81& \cellcolor{Gray0} \textbf{0.90}\\

0.97  & \cellcolor{Gray1}   0.99 \dag &\cellcolor{Gray1} 	0.98&\cellcolor{Gray1} 	0.98&	\cellcolor{Gray1} 0.96&\cellcolor{DarkGray}	1.08&\cellcolor{Gray1} 	0.96 &\cellcolor{DarkGray}	1.07 & \cellcolor{Gray1} \textbf{1.00} & \cellcolor{Gray1}   0.98&\cellcolor{Gray1} 	0.99  \dag&\cellcolor{Gray1} 	0.97& \cellcolor{Gray0}	0.93&\cellcolor{Gray1} 	0.97& \cellcolor{Gray0}	0.89&\cellcolor{DarkGray}	1.06& \cellcolor{Gray1} \textbf{0.97}\\

 0.99  & \cellcolor{DarkGray} 1.02&\cellcolor{Gray1} 	0.98  &\cellcolor{DarkGray}	1.01 \dag&\cellcolor{DarkGray}	1.10&\cellcolor{DarkGray}	1.21&\cellcolor{DarkGray}	1.23&\cellcolor{DarkGray}	1.20 & \cellcolor{DarkGray} \textbf{1.10} & \cellcolor{Gray1}  0.99  \dag &\cellcolor{Gray1} 	0.98& \cellcolor{Gray1}	1.00  &\cellcolor{Gray1}	1.00&\cellcolor{DarkGray}	1.22&\cellcolor{DarkGray}	1.18&\cellcolor{DarkGray}	1.10& \cellcolor{DarkGray} \textbf{1.06}\\

\hline
\textbf{Avg}  &  \cellcolor{Gray1}  \textbf{0.99}&\cellcolor{Gray1}  \textbf{0.97}&\cellcolor{Gray0}	 \textbf{0.91} & \cellcolor{Gray0}	 \textbf{0.82} & \cellcolor{Gray0}	 \textbf{0.82} & \cellcolor{Gray0}	 \textbf{0.81} &	 \textbf{0.69}  & \cellcolor{Gray0}\textbf{0.85}&\cellcolor{Gray1}    \textbf{0.99} &\cellcolor{Gray1} 	 \textbf{0.96} & \cellcolor{Gray0}	 \textbf{0.89} & \cellcolor{Gray0}	 \textbf{0.81} & \cellcolor{Gray0}	 \textbf{0.80} &	 \textbf{0.74} &	 \textbf{0.73} & \cellcolor{Gray0}  \textbf{0.84}\\
 \hline
\end{tabular}
\label{tab:performance_det1}
\end{table}
\renewcommand{\baselinestretch}{1.5} 

Table \ref{tab:performance_det2} extends the analysis to a larger number of machines and different processing time distributions. Each cell shows the average result for $u = \{0.8, 0.85, 0.9, 0.95, 0.97\}$, and all results are significant at the 0.05 level. As the number of machines increases, R1 and R2 tend to perform better than the benchmark rules (the value of $\tau$ decreases). The exception is the settings with very tight due dates ($a = 2$), where both rules deteriorate performance. We observed that as $|M|$ increases, the average queue size also increases, from ~4.5 (with ten machines) to ~7 (with 100 machines). This system behaviour affects the balance between WINQ and PT and is likely the reason for the worse performance of the rules in such settings. Another exception is $a = 8$ and $|M|$ = 100, where RR (the benchmark rule), R1, and R2 produce zero tardiness, resulting in $\tau = 1.00$. The processing time distribution has little effect on the performance of the rules.

\renewcommand{\baselinestretch}{1.3} 
 \begin{table}[H]
 \scalefont{0.8}
\caption{Performance for different number of machines and processing time distribution. }
\centering
\begin{tabular}{c c c c c c c c| c c c c c c}
\hline
& & \multicolumn{6}{c|}{\textbf{R1}} & \multicolumn{6}{c}{\textbf{R2}} \\
$\bar{p}$ & $|M| \ a$&  2 &   3&  4 &  6 &  8 & \textbf{Avg} &  2 &   3&  4 &  6 &  8 & \textbf{Avg} \\
\hline
\multirow{4}{*}{25 - 100} & 10 &  \cellcolor{Gray1} 0.97&\cellcolor{Gray0}	0.91&\cellcolor{Gray0}	0.84&	0.73&	0.72  & \cellcolor{Gray0} \textbf{0.83} & \cellcolor{Gray1} 0.97&\cellcolor{Gray0}	0.90&\cellcolor{Gray0}	0.83&	0.71&	0.67 & \cellcolor{Gray0} \textbf{0.81} \\

&  20 &\cellcolor{Gray1}   1.00& \cellcolor{Gray1} 	0.95&	0.79&	0.50&	0.38  & \textbf{0.68} & 1.00&\cellcolor{Gray0}	0.93&	0.77&	0.49&	0.36 & \textbf{0.66} \\

& 50 & \cellcolor{DarkGray} 1.06&\cellcolor{Gray0}	0.88&	0.33&	0.41&	0.40  & \textbf{0.55} & \cellcolor{DarkGray}1.04&\cellcolor{Gray0}	0.88&	0.36&	0.45&	0.40 & \textbf{0.57} \\

& 100 & \cellcolor{DarkGray} 1.08&	0.59&	0.17&	0.18&\cellcolor{Gray1} 	1.00& \textbf{0.45} &  \cellcolor{DarkGray} 1.06&	0.64&	0.33&	0.33&\cellcolor{Gray1} 	1.00 & \textbf{0.59}
\\

 \hline
250 & 10 & \cellcolor{Gray1}   0.96&\cellcolor{Gray0}	0.91&\cellcolor{Gray0}	0.84&	0.73&	0.76 & \cellcolor{Gray0} \textbf{0.84} & \cellcolor{Gray1}   0.96&	\cellcolor{Gray0}0.90	&\cellcolor{Gray0} 0.82	&0.73&	0.70 & \cellcolor{Gray0} \textbf{0.82} \\

 \hline
500 & 10 &  \cellcolor{Gray1} 0.96&\cellcolor{Gray0}	0.92&\cellcolor{Gray0}	0.85&	0.72&	0.72 & \cellcolor{Gray0} \textbf{0.83} & \cellcolor{Gray1}   0.96&\cellcolor{Gray0}	0.89&\cellcolor{Gray0}	0.83&	0.71&	0.67 & \cellcolor{Gray0} \textbf{0.80}\\

 \hline
\end{tabular}
\label{tab:performance_det2}
\end{table}
\renewcommand{\baselinestretch}{1.5} 

\subsection{Performance under stochastic processing times}

To assess the effectiveness of the evolved rules under uncertain processing times, we compare the mean tardiness generated by each rule for different \textit{cv} values. As explained in Section \ref{S:instances}, \textit{cv} is the coefficient of variation for sampling the actual processing times, (larger values for \textit{cv} correspond to higher uncertainty levels; when \textit{cv} = 0, processing times are deterministic). In Table \ref{tab:perf_stc2}, we report the average value of $\tau$ under stochastic processing times. Although the relative performance of both rules deteriorates, they are still better than the benchmark -- notice that the training set is composed of instances with deterministic processing times. The average performance of R1 and R2 for each setting is presented in Section \ref{s:performance_stc} of Appendix.

\renewcommand{\baselinestretch}{1.3} 
\begin{table}[ht]
 \scalefont{0.8}
\caption{Overall performance under stochastic processing times}
\centering
\begin{tabular}{l c c c c c c c}
\hline
\multirow{2}{*}{\textbf{Rule}} & \multicolumn{6}{c}{ \textbf{\textit{cv}}} & \multirow{2}{*}{\textbf{Avg}} \\
 & 0.0 & 0.2 & 0.4 & 0.6 & 0.8 & 1.0 &  \\
\hline
R1 & \cellcolor{Gray0} 0.83 &\cellcolor{Gray0}	0.91 &\cellcolor{Gray0}	0.92 &	\cellcolor{Gray0}0.90 &\cellcolor{Gray0}	0.92 &\cellcolor{Gray0}	0.91 &\cellcolor{Gray0}	0.90  \\
R2 & \cellcolor{Gray0} 0.81 &\cellcolor{Gray0}	0.85 &\cellcolor{Gray0}	0.86 &\cellcolor{Gray0}	0.87 &\cellcolor{Gray0}	0.88 &\cellcolor{Gray0}	0.89 &\cellcolor{Gray0}	0.86  \\
\hline
\end{tabular}
\label{tab:perf_stc2}
\end{table}
\renewcommand{\baselinestretch}{1.5} 

\begin{figure*}[ht]
\begin{center}
\begin{subfigure}{0.33\textwidth}
	\includegraphics[width=\textwidth]{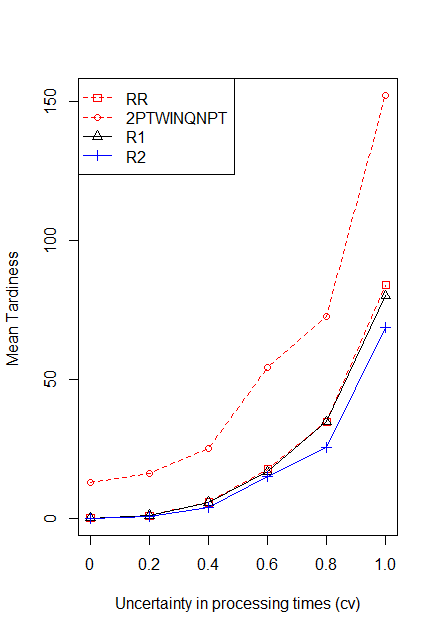}
	\caption{Light load conditions, $a$ = 8 and $u$ = 0.89} \label{fig:8a}
\end{subfigure}
\begin{subfigure}{0.33\textwidth}
	\includegraphics[width=\textwidth]{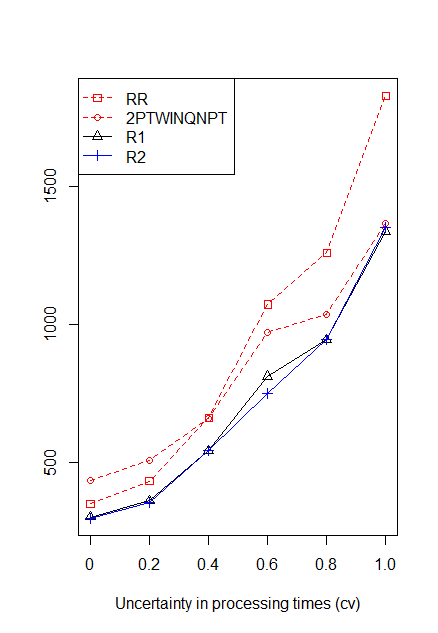}
	\caption{Medium load conditions, $a$ = 4 and $u$ = 0.90} \label{fig:8b}
\end{subfigure}	
\begin{subfigure}{0.33\textwidth}
	\includegraphics[width=\textwidth]{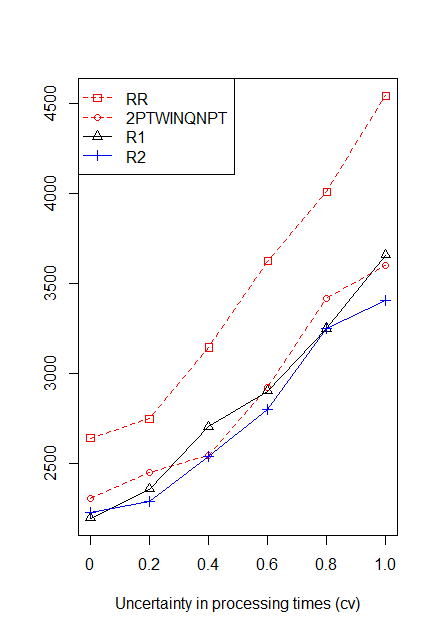}
	\caption{Tight load conditions, $a$ = 2 and $u$ = 0.97} \label{fig:8c}
\end{subfigure}	
\caption{Mean tardiness generated by evolved and existing rules for stochastic processing times.} \label{fig:tardiness_stc}
\end{center}
\end{figure*}

Our analysis finishes with the effect of uncertainty in the mean tardiness for distinct load conditions. The charts in Figure \ref{fig:tardiness_stc} present the mean tardiness $\bar{T}$ obtained for rules RR, 2PT+WINQ+NPT, R1 and R2, considering all uncertainty levels. Notice that now we report absolute values (unlike the previous sections). These results refer to instances where $\bar{p} = 50$ and $|M| = 10$. In general, all dispatching rules produce higher tardiness as \textit{cv} increases. Comparing the existing rules, RR and 2PT+WINQ+NPT are good in light and tight load conditions, respectively, but none of them works well in both cases. In medium conditions (Figure \ref{fig:8b}), RR performs well if the \textit{cv} is low; as uncertainty increases its performance degrades fast. R2 consistently produces lower tardiness than both, regardless of the setting, and R1 is good in medium conditions. 

\section{Conclusions and future work}    
\label{S:conclusions}

This work focuses on the use of problem expertise to leverage the computational power of machine learning. Genetic Programming is applied as the learning approach on the design of dispatching rules for the Dynamic Job Shop Scheduling Problem. The objective is to find rules that minimise mean tardiness, one of the most challenging performance criteria in dynamic job shops. We propose a guided empirical learning procedure, where reasoning on the problem derives insights to guide the algorithmic search in an iterative way. At each iteration, modifications are applied to the terminal set or rules structure, and new results are generated and analysed. Unlike traditional approaches, GP is also provided with problem expertise from the existing literature.

We found that GP can combine parameters from existing rules and generate better ones. Another finding is that a small terminal set of selected parameters enables evolving more effective dispatching rules than working with large terminal sets. Moreover, we conclude that the overall effectiveness of the evolved rules is even better when providing a fixed good structure. 

All rules generated were evaluated in two criteria, performance and size, and the non-dominated were selected for testing. Besides having an interpretable structure, these rules present an overall improvement of 19\% compared to the best possible combination of the state-of-the-art rules. This superiority is achieved in almost all settings, from extreme light load conditions (where all jobs meet the due date) to congested ones. Among all tested settings, only for extremely tight conditions, where shop utilisation is around 99\%, the average performance of the new rules does not outperform the literature. Furthermore, the good behaviour of the rules is confirmed in shops with a different number of machines and under stochastic processing times. 

As future work, our guided empirical learning approach should be tested in different problems, such as the Flexible Job Shop Scheduling or Parallel Machine Scheduling. Also, as there are several performance criteria in all these scheduling problems, multi-objective optimisation becomes a relevant research topic. 

\section*{Acknowledgments}

This work is financed by the ERDF – European Regional Development Fund through the Operational Programme for Competitiveness and Internationalisation – COMPETE 2020 Programme, and by National Funds through the Portuguese funding agency, FCT - Funda\c{c}\~{a}o para a Ci\^{e}ncia e a Tecnologia, within project SAICTPAC/0034/2015- POCI-01-0145-FEDER-016418

\bibliographystyle{elsarticle-harv}
\bibliography{references}

\begin{appendices}
\label{S:appendix}

\section{Job Shop Simulator}
\label{S:simulator_alg}

The pseudo-code of the job shop simulator is described in Algorithm \ref{alg:simulator}. The inputs are instance parameters <$|M|$, $\bar{p}$, $a$, $u$, \textit{cv}> and a dispatching rule. As described in Section \ref{S:simulator}, each job arrival $r_j$ is sampled from a Poisson distribution which mean rate depends on the value of $u$. The allowance factor $a$ is considered to calculate jobs' due dates. The parameter $\bar{p}$ is the mean processing time, and \textit{cv} is the coefficient of variation for sampling the actual processing times. Table \ref{tab:variables} describes some of the variables defined in the algorithm and the respective initialisation. The next operation of each job enters machines' queue $Q_m$ at time $start_{j}$, when they are available for execution.

\begin{table}[H]
\caption{Main variables considered by Algorithm \ref{alg:simulator}}
\centering
    \begin{tabular}{l l l}
        Variable & Description & Initialisation  \\
       \hline
        $num\_jobs$ & Number of finished jobs  & 0\\
        $J'$ & Jobs not completed  & $J$\\
        $start_{j}$ & Minimum start time of job $j$ next operation  & $r_j$\\
        $m\_next_j$ & Next machine of job $j \in J$  & First machine of $j$\\
        $Q_m$ & Queue in machine $m$  & $\emptyset$\\
        $A_m$ & Next available time of machine $m$  & $0$\\
\hline
    \end{tabular}
     \label{tab:variables}
\end{table}

The simulation starts with an empty shopfloor and goes on until a number of jobs have been computed (max\_jobs). The next operation of each job enters the machines' queue ($Q_m$) at time $start_{j}$, when they are available for execution. In line 2, the algorithm verifies which is the earliest event between the availability of a new operation of a job ($start_j$) and the availability of machine ($A_m$). An operation becomes available upon job arrival or when the predecessor operation finishes. In both cases, the operation may enter the next machines' queue (line 4), and the value of $start_j$ is set to $\infty$. Once a machine whose queue is not empty becomes available (line 7), the next operation to be executed is selected. In lines 8 to 10, the dispatching rule runs and calculates a score for each job in the queue. The job $j*$ with the least score is executed (in case of ties, the algorithm selects the job that arrived earlier to the queue). If $j*$ has no operations left, it is completed, removed from $J'$ and the tardiness is calculated (lines 15-18). Otherwise, the algorithm updates the start time of the jobs' next operation (line 20). We recall that jobs are numbered on arrival in a crescent sequence order, and the tardiness $T_j$ is computed only for jobs numbered from $min$ to $max$. Following the values adopted in literature, we set $min = 501$ and $max = 2500$, max\_jobs = $max - min$ = 2000.

\begin{algorithm}[H]
\SetAlgoLined
\SetKwInOut{Input}{Input}
\Input{Instance <$|M|$, $\bar{p}$, $a$, $u$, cv> and DR}
\KwResult{Mean Tardiness $\bar{T}$}
\While{$num\_jobs < max\_jobs$}{
\eIf{$\min\limits_{j \in J'} \{start_j\}$ < $\min\limits_{m \in M: Q_m \neq \emptyset}\{A_m\}$}{
   \emph{$j^{*} \leftarrow \argmin\limits_{j \in J'} \{start_j\}$} \;
    \emph{$Q_{m\_next_{j^{*}}} \leftarrow Q_{m\_next_{j^{*}}} + j^{*}$}\;
    \emph{$start_{j^{*}} \leftarrow  \infty$}\;
}{
    \emph{$m \leftarrow \argmin\limits_{m \in M: Q_m \neq \emptyset} \{A_m\}$}\;
        
    \ForEach{job $j \in Q_m$}{
    \emph{$score_j \leftarrow$ run\_DR on j}\;
    }
    \emph{$j^{*} \leftarrow \argmin\limits_{j \in Q_m}\{score_j\}$}\;
    \emph{execute j* in m}\;
    \emph{$A_m \leftarrow  finish\_time_{j^{*}} $}\;
   
        \eIf{$m$ = last\_machine of $j^{*}$}{
        \emph{$J' \leftarrow J' \setminus j^{*}$}\;         
            \If{$j^{*} \in \{min...max\}$}{
                \emph{$num\_jobs \leftarrow num\_jobs +1$}\;
                  \emph{$T_{j^{*}} \leftarrow max(0, d_{j^{*}} - finish\_time_{j^{*}})$}\;
                  \emph{$\bar{T} \leftarrow \bar{T} + T_{j^{*}}$ }\;
            }
        }{
        \emph{$start_{j^{*}} \leftarrow  finish\_time_{j^{*}}$} \;
        }
}
}
\emph{$\bar{T} \leftarrow \bar{T} \div $ max\_jobs}\;
\caption{Job Shop Simulator}\label{alg:simulator}
\end{algorithm}
 
\section{Terminal sets considered in GP experiments}
\label{app:terminals}

Table \ref{tab:terminalsGP1} lists all terminals used in the first learning iteration, where two distinct sets are proposed. For calculating the evolved rules' size, we assume that all terminals in these sets have size one. 

\begin{table}[H]
\caption{Terminal sets for GPLit}
\centering
\begin{tabular}{l c c }
\hline
\textbf{Description} & \textbf{GPLit$_a$}  & \textbf{GPLit$_b$} \\
\hline
Average processing time of operations in current queue & x &x  \\
\hline
Expected processing time of current operation & x  & x \\
\hline
Job allowance&   &x  \\
\hline
 Job arrival time& x  &  \\
\hline
Job due date& x  &  \\
\hline
Job queuing time&   &  x\\
\hline
Machine ready time& x  & x \\
\hline
Maximum due date time among jobs in queue&  & x \\
\hline
 Maximum processing time among operations in queue&   & x \\
\hline
Minimum due date time among jobs in queue&  & x \\
\hline
Minimum processing time among operations in queue &   & x \\
\hline
Number of jobs in the system &  & x \\
\hline
Number of operations left & x& x \\
\hline
Operation ready time & x  &  \\
\hline
Processing time of the next operation of the job & x  & x  \\
\hline
Queue size in current machine &   & x  \\
\hline
Queue size in the next machine &  & x \\
\hline
Remaining  processing time of job & x & x \\
\hline
Remaining processing time of all jobs in the queue &   & x \\
\hline
Remaining time of the successive operations&  &  \\
\hline
Slack of job & x   &x  \\
\hline
 Time in the system& x & x \\
\hline
 Work in current machines' queue &  &  \\
\hline
Work in the next machines' queue &  x  & x \\
\hline
 Work in all successive machines' queues &   & x \\
\hline

\end{tabular}
\label{tab:terminalsGP1}
\end{table}

\section{Performance of evolved rules under stochastic processing times}
\label{s:performance_stc}

Table \ref{tab:performance_stc1} presents the performance $\tau$ of R1 and R2 for each combination of allowance factor and shop utilisation. Despite some variance in the results, the relative performance of both rules tends to be more homogeneous compared to settings with deterministic processing times. Moreover, both rules behaviour is better under light load conditions.

  \renewcommand{\baselinestretch}{1.3} 
 \begin{table}[H]
 \scalefont{0.8}
\caption{Performance under stochastic processing times for each setting.}
\centering
\begin{tabular}{c c c c c c c   | c c c c c c }
\hline
 & \multicolumn{6}{c}{\textbf{R1}} & \multicolumn{6}{|c}{\textbf{R2}} \\
$u \ a$& 2 &  3& 4 & 6 & 8 & \textbf{Avg} & 2 & 3& 4 & 6 & 8  & \textbf{Avg}\\
\hline
0.80 & \cellcolor{Gray1}0.95	& \cellcolor{Gray0}  0.89&	\cellcolor{Gray0}0.83&	0.78&	\cellcolor{Gray1}0.97&	\cellcolor{Gray0}\textbf{0.88}& \cellcolor{Gray1}0.95&\cellcolor{Gray0}	0.88&\cellcolor{Gray0}	0.82&	0.68&	0.76&	\cellcolor{Gray0}\textbf{0.81}\\

0.85 & \cellcolor{Gray1}0.97&\cellcolor{Gray0}	0.89&\cellcolor{Gray0}	0.85&	0.77&	0.77&\cellcolor{Gray0}	\textbf{0.84}& \cellcolor{Gray1}0.96&\cellcolor{Gray0}	0.88&\cellcolor{Gray0}	0.83&	0.74&	0.62&\cellcolor{Gray0}	\textbf{0.80}\\

0.90  & \cellcolor{Gray1}0.96&\cellcolor{Gray1}	0.93	&\cellcolor{Gray0}0.88&	\cellcolor{Gray0}0.82&	0.78&	\cellcolor{Gray0}\textbf{0.87}& \cellcolor{Gray1}0.95&\cellcolor{Gray0}	0.91&\cellcolor{Gray0}	0.86&	0.79&	0.73&	\cellcolor{Gray0}\textbf{0.85}\\

0.95  & \cellcolor{Gray1}0.97&	\cellcolor{Gray1}0.97&\cellcolor{Gray0}	0.93&\cellcolor{Gray0}	0.89&\cellcolor{Gray0}	0.88&\cellcolor{Gray0}	\textbf{0.93}& \cellcolor{Gray1}0.96&	\cellcolor{Gray1}0.96&\cellcolor{Gray0}	0.92&\cellcolor{Gray0}	0.86&\cellcolor{Gray0}	0.82&\cellcolor{Gray0}	\textbf{0.90}\\

0.97  & \cellcolor{Gray1}0.98&\cellcolor{Gray1}	0.97&\cellcolor{Gray1}	0.97&\cellcolor{Gray1}	0.96&\cellcolor{Gray0}	0.94&\cellcolor{Gray1}	\textbf{0.97}& \cellcolor{Gray1}0.97&\cellcolor{Gray1} 0.97&\cellcolor{Gray0}	0.93&\cellcolor{Gray0}	0.93&\cellcolor{Gray0}	0.91&\cellcolor{Gray0}	\textbf{0.94}\\

\hline
\textbf{Avg}  &\cellcolor{Gray1} \textbf{0.97}&	\cellcolor{Gray0}\textbf{0.93}	&\cellcolor{Gray0}\textbf{0.89}&	\cellcolor{Gray0}\textbf{0.84}&\cellcolor{Gray0}	\textbf{0.86}&	\cellcolor{Gray0}\textbf{0.90}& \cellcolor{Gray1} \textbf{0.96}&\cellcolor{Gray0}	\textbf{0.92}&\cellcolor{Gray0}	\textbf{0.87}&\cellcolor{Gray0}	\textbf{0.80}&	\textbf{0.76}&\cellcolor{Gray0}	\textbf{0.86}\\

 \hline
\end{tabular}
\label{tab:performance_stc1}
\end{table}
\renewcommand{\baselinestretch}{1.5} 

\end{appendices}

\end{document}